# Simulación de la distribución de alimento en el cultivo de camarón

1st Renato L. Conforme Rosado
*Facultad de Ingenieria*
*Pontificia Universidad Javeriana*
Bogota, Colombia
0000-0003-1966-271X
renato_conforme@javeriana.edu.co

2nd Francisco C. Calderon Bocanegra, PhD
*Facultad de Ingenieria*
*Pontificia Universidad Javeriana*
Bogota, Colombia
0000-0001-8681-415X
calderonf@javeriana.edu.co

*Abstract*— This document presents the experimentation of 4 cases of food distribution for shrimp farming. The distributions are based on the location of the automatic feeders. Three cases applied in reality and a fourth case where the food is irrigated on the crop simultaneously and uniformly. In a first stage, the simulation of the three distribution cases is successfully adjusted to reality, where the trend of the shrimp growth curve is correlated with the historical data curve. A second stage where you experiment in 16 configurations that are based on the amount of food, the density of biomass and the distribution of the food. The simulation adopts the concepts of genetic algorithms to improve the population and fuzzy logic as an agent evaluation technique for decision-making against the quality of physical-chemical parameters in the simulated environment. The results of these interactions reveal a reduction in the simulated total culture time from 22 weeks to 14 weeks.

*Keywords—simulation, multi-agents, cellular automata, shrimp culture.*

*Resumen*—Este documento presenta la experimentación de 4 casos de distribución del alimento para el cultivo de camarón. Las distribuciones están en función de la ubicación de los alimentadores automáticos. Tres casos aplicados en la realidad y un cuarto caso donde se dispone el riego del alimento sobre el cultivo de manera simultánea y uniforme. En una primera etapa se ajusta con éxito la simulación de los tres casos de distribución a la realidad, donde la tendencia de la curva de crecimiento del camarón esta correlacionada con la curva de datos históricos. Una segunda etapa donde se experimenta en 16 configuraciones que están en función de la cantidad de alimento, la densidad de biomasa y la distribución del alimento. La simulación adopta los conceptos de algoritmos genéticos para mejorar la población y lógica difusa como técnica de evaluación del agente para la toma de decisiones frente a la calidad de parámetros físicos-químicos en el ambiente simulado. Los resultados de estas interacciones revelan una reducción en el tiempo simulado del cultivo total de 22 semanas a 14 semanas.

*Palabras claves—simulación, multi-agentes, autómatas celulares, cultivo de camarón.*

## I. Introducción

La acuicultura presentó un crecimiento entre los años 1985 y 2002 de 7,603 por ciento, mostrándose como una actividad económica promisoria y altamente contribuyente hacia el futuro de la producción total colombiana. El cultivo camaronero es un área representativa de la acuicultura y en países vecinos como Ecuador su producción alcanza las 120.000 toneladas con 180.000 hectáreas para el cultivo, en Colombia con 2.000 hectáreas produce 8.000 toneladas incluida la pesca de captura [1].

Uno de los procesos que más gastos representa en la producción total es la alimentación, el alimento del animal llega hasta un 30% del costo total de una cosecha. El riego excesivo del alimento afecta a la calidad del agua, lo que a su vez está afectando al desarrollo del animal, de la misma forma perjudica si el riego es inferior. Entonces cualquier modificación que se realice en algún proceso del cultivo, afectará positiva o negativamente a la cosecha.

La existencia de una herramienta que ayude a mitigar los riesgos involucrados en la producción sería de considerable utilidad, ya que sus consecuencias no son medidas en la realidad sino en un ambiente simulado.

## II. Descripción

Los conceptos de inteligencia artificial que mejor se adaptan a este trabajo son: Lógica difusa, autómatas celulares con "flocking" y algoritmos genéticos. La combinación de estas técnicas, es decir, sistemas híbridos, se integran en un solo sistema [2] [3].

El desarrollo de cada agente y sus características serán implementadas bajo el concepto de "flocking" [4] [5] [6] [7]. La implementación de los agentes será independiente uno del otro, lo que le permite ser replicados las veces que sean necesarias, como lo es en el caso del agente camarón.

El presente trabajo está desarrollado en un solo contenedor, es decir, no será un sistema distribuido aplicando sistemas multi-agentes [8]. Los agentes cobran vida sobre una plataforma que le proporciona servicios de entrega de mensajes. La plataforma está compuesta de un contenedor, el agente hábitat hará las veces de contenedor, quiere decir que es el responsable de la comunicación entre los agentes.

La simulación inicia con una muestra de la población real, ya que el costo computacional es muy alto. Esto quiere decir que todo el sistema está en una escala de 1:8000. La población es homogénea inicialmente, es decir, todos los agentes camarón tienen el mismo código genético y a medida que avanza en las generaciones, esta población se va diversificando mediante el uso de un algoritmo genético. En el modelo del algoritmo genético, los agentes interactúan usando tres reglas principales: Alimentarse, explorar y tolerancia al ambiente, esta última es evaluada por lógica difusa.

Al final de cada época se cuestiona si se alcanzó el promedio de biomasa esperado, si es así, se acumula los datos de esa última generación, hasta que se cumpla el máximo de generaciones. Si se da el caso contrario y no se cumple con el número de generaciones esperado, se hereda a una nueva generación las características de la población con mejor rendimiento y tolerancia al ambiente. El proceso se repetirá hasta que se alcance el número de generaciones con el promedio de biomasa esperado. Finalmente se elige la generación con menor tasa de error como se muestra en la Figura 15.

A. *Aplicando algoritmos geneticos*

Este apartado describe la capa externa de la simulación que es la adaptación de un algoritmo genético para elegir una población diversificada, que permita a la simulación ser un reflejo de la realidad. Si se eligiera que todos los agentes camarones estén en perfectas condiciones, no sería una realidad ya que los históricos de una producción revelan que no toda la población tiene un tamaño estándar sino distintos grupos en distintos tamaños. Por otro lado, si se elige una población aleatoria, para el final de cada simulación la curva de resultados será totalmente distinta, y tendría datos incoherentes y menos confiables para una comparación con una curva real en el cultivo.

*1) Cromosoma*

El cromosoma es una representación de la especie, la cual contiene una cadena de valores cuantitativos de las características, propiedades, atributos, habilidades o la combinación de todas las anteriores para dicha especie. Para la representación del camarón en un cromosoma se combinó la tolerancia al ambiente, habilidades y atributos del camarón. Como se muestra en la Figura 1, el cromosoma está compuesto por dos cadenas de valores, la tolerancia al ambiente y las propiedades.

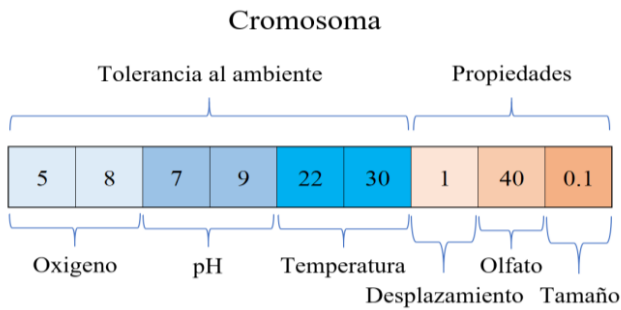

*Figura 1. Estructura del cromosoma para el agente camarón*

La primera cadena está formada por los parámetros físicos-químicos. Que son: oxígeno, pH y temperatura. Cada parámetro posee dos campos, un mínimo y un máximo tolerados por la especie ubicados de izquierda a derecha respectivamente. La diferencia entre estos campos da el valor de tolerancia a la variación de los parámetros en el ambiente simulado.

La segunda cadena está compuesta por el desplazamiento, olfato y tamaño, estas propiedades no serán sometidas a modificación genética en el método de reproducción explicado más adelante. Esta se trata de las ganancias obtenidas por los camarones en la simulación o dicho de otra manera, su desarrollo; su principal aportación es la capacidad de adaptación al ambiente simulado.

*2) Método de selección*

Un algoritmo genético puede utilizar muchas técnicas diferentes para seleccionar a los individuos que deben copiarse hacia la siguiente generación. El principal valor para la elección de estos individuos es la función de evaluación o "fitness", esta es una representación cuantitativa que le aporta valor y permite la evaluación de dicho cromosoma.

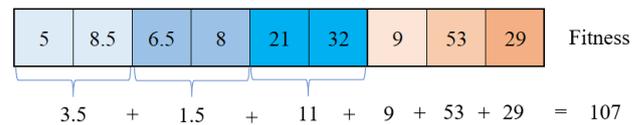

*Figura 2. Cálculo de la función de evaluación*

Para obtener la función de evaluación primero se obtiene las diferencias entre los máximos y mínimos de cada parámetro físico - químico luego se suma con los valores de las propiedades y con este valor el cromosoma se somete a selección. Un ejemplo de esto puede verse en la Figura 2.

Para este trabajo se eligió el método de selección por torneo, este consiste en hacer subgrupos al azar de individuos de la población como se muestra en la Figura 3; se elige los miembros de mayor fitness de cada subgrupo y estos son denominados padres para la reproducción.

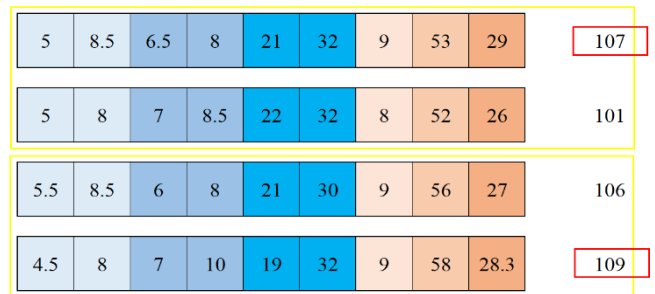

*Figura 3. Selección de los subgrupos*

*3) Método de reproducción*

En este método los cromosomas padres son sometidos a un proceso de mejoramiento como lo es el cruzamiento, este consiste en seleccionar a dos individuos para que intercambien segmentos de su código genético o cromosoma, produciendo una "descendencia" artificial cuyos individuos son combinaciones de sus padres.

Esta combinación solo ocurre en el primer segmento del cromosoma y de a pares, respetando la secuencia de cada parámetro tal como se muestra en la Figura 4. Este método es el que mejor se adapta a la aplicación, ya que se toma los mejores valores de cada padre para la combinación de estos parámetros y hace posible la reproducción al tratarse de cromosomas de la misma especie.

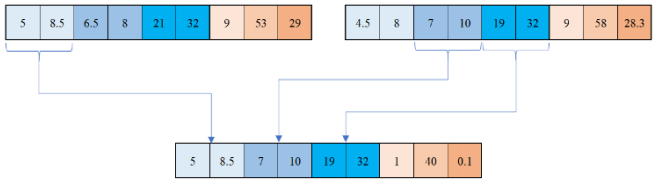

*Figura 4. Descripción del cruce de cromosomas*

### B. Automatas celulares

Para lograr tener una interacción en la simulación los agentes poseen unas reglas condicionales que resulta en un comportamiento masivo emergente. Los agentes camarones como autómatas celulares tienen la capacidad de elegir entre alimentarse, explorar y la tolerancia al ambiente. El agente camarón tiene dos movimientos. El movimiento exploratorio donde el realiza movimientos en direcciones aleatorias frente al estado del ambiente, el cual influye sobre la dirección que va a tomar. El segundo movimiento es el de cacería; es cuando el agente camarón detecta comida por medio del olfato, en este caso ignora las condiciones del ambiente para moverse hacia el agente alimento y alimentarse.

*1) Ambiente*

El estado del ambiente está en función del número de agentes camarones por área, es decir, la densidad, a mayor densidad menor la calidad del ambiente en dicha área. Se dividió en cuatro niveles de calidad: Bueno, medio, tolerable y malo. Para la visualización en la simulación se los representa con los colores: Blanco, verde, amarillo y rojo respectivamente, como se presenta en la Figura 5.

Cuando la condición del ambiente varía, específicamente cuando su calidad baja, éste afecta a la dirección del movimiento exploratorio del agente camarón; cabe recalcar que dicho movimiento es posible por la no existencia del agente alimento.

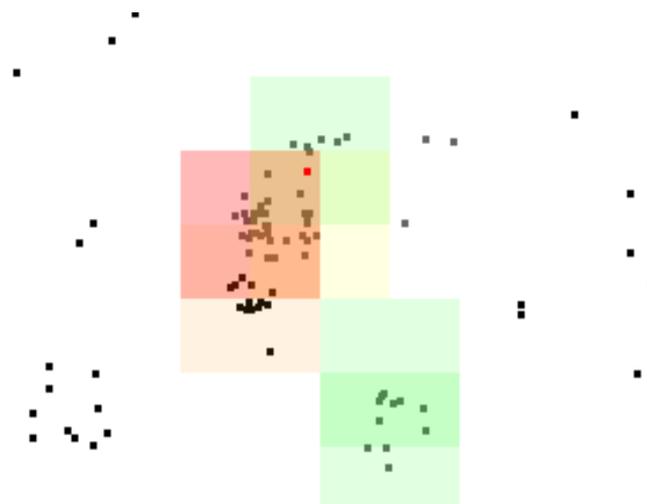

*Figura 5. Fragmento de la simulación, se observa que el área más congestionada esta de rojo y el área menos congestionada esta de blanco*

*2) Alimento*

El agente alimento como tal, es un agente estático que no interactúa con el ambiente, pero lo afecta indirectamente, ya que su presencia ocasiona la aglomeración por parte del agente camarón, en consecuencia, los niveles de calidad del ambiente bajan.

El alimento se riega periódicamente y está distribuido en tres zonas de la piscina. Se lo realiza de esta manera imitando las técnicas más comunes de riego actuales. En condiciones experimentales de la simulación, el riego del alimento es distribuido de diferentes maneras para comprobar su eficiencia en el desarrollo del agente camarón.

*3) Camarón*

El agente camarón, como se mencionó con anterioridad, tiene dos estados, uno pasivo y otro activo, estos estados se refieren a la condición de movilidad. El estado activo depende directamente de su capacidad de olfato, quiere decir que puede detectar un alimento y una vez detectado se mueve hacia él, caso contrario, su estado es pasivo y su movimiento es exploratorio.

Además, el agente camarón es afectado por el ambiente. Cuando la calidad del ambiente baja, se reduce la capacidad de desplazamiento y olfato, cumpliendo con la representación del estrés del camarón en la simulación. Específicamente, cuando al camarón le afectan las condiciones ambientales y se reducen las oportunidades de alimentarse.

*4) Simulación del olfato*

El estado activo del agente camarón es consecuente de la propiedad de olfato, en otras palabras, que el camarón detecte el alimento más cercano para moverse hacia él. El olfato es simulado con la distancia mínima entre dos puntos. Cada agente camarón tiene un radio de alcance para su olfato y calcula la distancia entre todas las comidas detectadas dentro de su radio de cobertura, donde D es la distancia entre el agente alimento y el agente camarón (Ver Figura 6); finalmente elige la comida con la menor distancia.

$$D = \sqrt{(lat\_alim - lat\_cam)^2 + (lon\_alim - lon\_cam)^2}$$

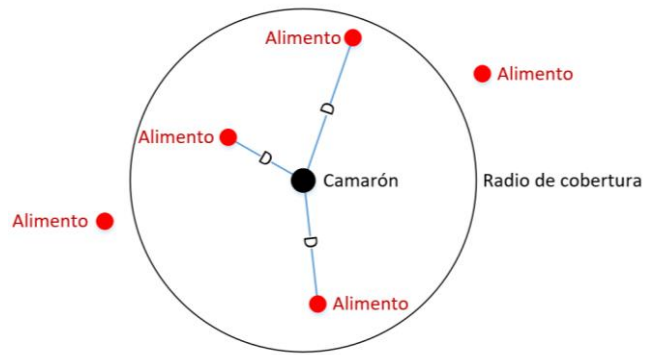

*Figura 6. Simulación del olfato, elige el alimento con menor distancia*

Una vez el agente camarón eligió el alimento a cazar, se desplaza hacia él, restando o sumando su posición actual en cada una de las coordenadas según la orientación hasta que su posición sea la misma del agente alimento (Ver Figura 7); es aquí cuando se suprime la existencia del agente alimento y se suma el contador de alimentación para el agente camarón.

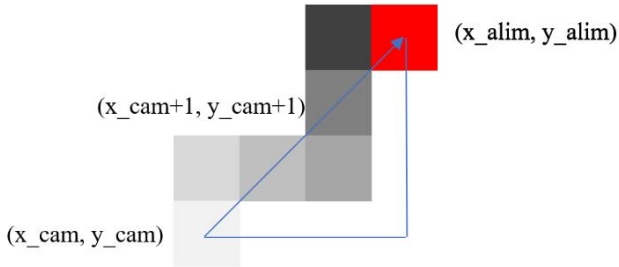

*Figura 7. Descripción del movimiento del agente camarón hacia el agente alimento*

5) *Movimiento exploratorio*

Para el agente camarón el movimiento exploratorio es donde realiza movimientos en direcciones aleatorias, si el estado del ambiente cambia a parámetros no tolerados por el agente camarón, este modifica su sentido en dirección opuesta a la zona afectada.

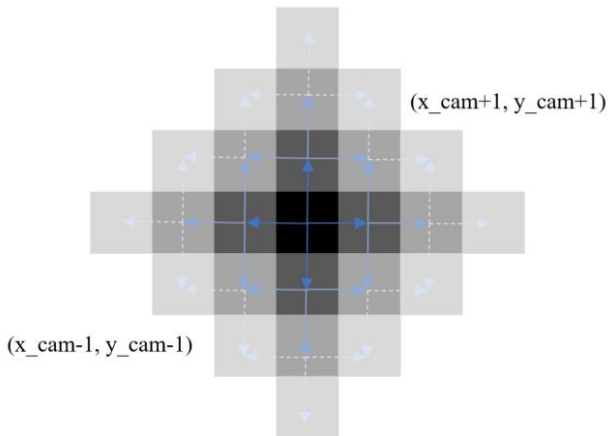

*Figura 8. Descripción del movimiento exploratorio del agente camarón*

El movimiento exploratorio es el cambio de posición del agente camarón en sentido horizontal o vertical pero no en diagonal. Por medio de una función aleatoria se elige en qué dirección se va a mover, sea derecha, izquierda, arriba o debajo de la posición inicial, siempre se mueve en una unidad de espacio (Ver Figura 8).

C. *Aplicando lógica difusa*

La afectación de los parámetros físicos-químicos sobre la especie se define mediante lógica difusa, ya que esta técnica permite discretizar valores entre un estado y otro.

Para el caso del agente camarón se definen tres estados: Normal (N), tolerable (T) y malo (M), existe un cuarto estado que no se está representando en la simulación, la muerte. Con respecto a la tolerancia del agente camarón a los parámetros fisicoquímicos, se determina como optima tolerancia y mala tolerancia. Ya definido los estados entonces se pueden crear las reglas para el proceso de fusificación. (Ver Tabla 1)

| Estado | Oxi. | pH | Temp. |
|--------|------|------|--------|
| N | Optimo | Optimo | Optimo |
| T | Optimo | Optimo | Malo |
| T | Optimo | Malo | Optimo |
| T | Malo | Optimo | Optimo |
| M | Malo | Malo | Optimo |
| M | Malo | Optimo | Malo |
| M | Optimo | Malo | Malo |
| - | Malo | Malo | Malo |

*Tabla 1. Condiciones de los parámetros para cada uno de los estados del agente camarón*

El área de evaluación es la misma área de la simulación y ésta consiste en una rejilla donde cada cuadro contiene información de los tres parámetros fisicoquímicos. El área de evaluación es afectada por la presencia de los agentes camarón y alimento. De acuerdo con lo mencionado antes sobre como los parámetros afectan en conjunto, los tres parámetros son evaluados con la operación difusa de la intersección. Para cada parámetro se escoge un criterio con la finalidad de diferenciarlos o clasificarlos.

1) *Conjuntos difusos*

Cada universo tiene dos conjuntos de pertenencia de acuerdo con la Tabla 2 son optimo (o) y malo (m). Entonces el universo de los conjuntos difusos (Ver Figura 22) de cada parámetro ambiental se establece de la siguiente manera:

- Oxígeno disuelto, su universo son los números reales entre 0 y 14 partes por millón "ppm".
  $A = \{X \in R \mid X = [0 - 14]\}$
  $A_o = \{x \in [x \geq 5 \wedge x \leq 12]\}$
  $A_m = \{x \in [x < 5 \wedge x > 12]\}$

- Potencial de Hidrógeno o pH, su universo son los números reales entre 0 y 14 dados por el logaritmo negativo de la actividad de los iones de hidrógeno.
  $B = \{Y \in R \mid Y = [0 - 14]\}$
  $B_o = \{y \in [y \geq 6.5 \wedge y \leq 8.5]\}$
  $B_m = \{y \in [y < 6.5 \wedge y > 8.5]\}$

- Temperatura, su universo son los números reales entre 18 y 36 grados Celsius.
  $C = \{Z \in R \mid Z = [18 - 36]\}$
  $C_o = \{x \in [x \geq 22 \wedge x \leq 30]\}$
  $C_m = \{x \in [x < 22 \wedge x > 30]\}$

## 2) Función de membresía

La función de membresía representa al conjunto difuso y su notación es μ(x), μ(y), μ(z) para sus conjuntos difusos respectivamente (Ver Figura 9). Sus dominios son; x ∈ X, y ∈ Y, z ∈ Z y para el rango está definido entre los valores 0 a 1, es decir; μ ∈ [0 - 1], donde 0 quiere decir que no hay pertenencia del elemento al conjunto y 1 que hay total pertenencia.

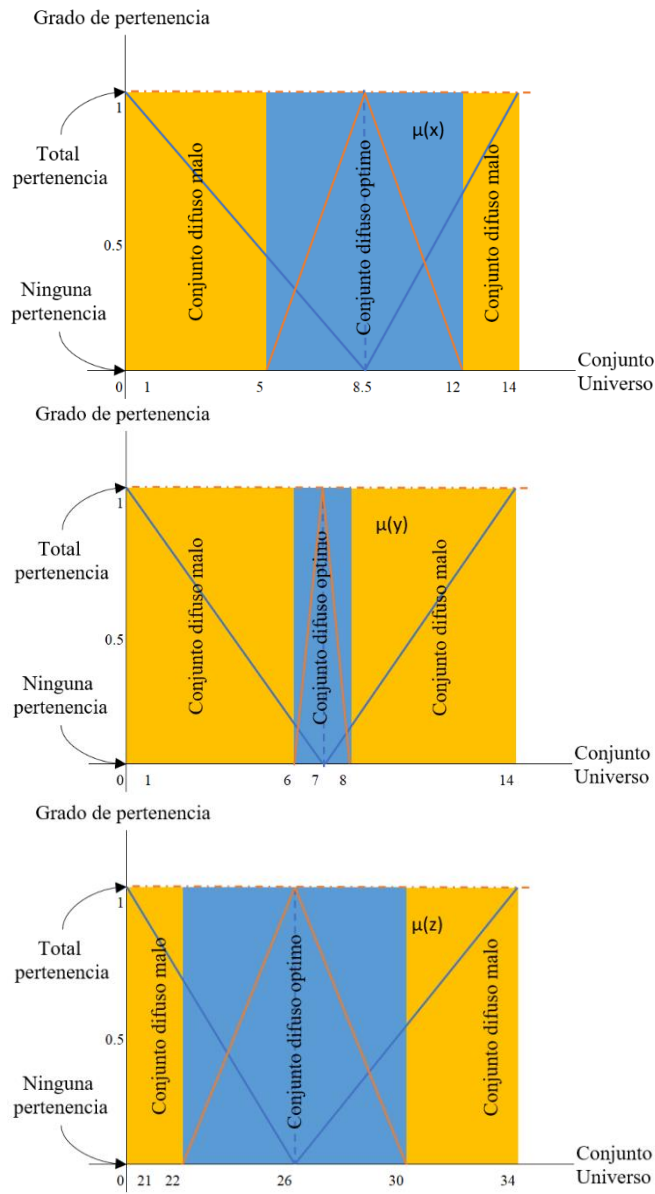

*Figura 9. Funciones de membresía de los parámetros, oxígeno disuelto, potencial de hidrogeno y temperatura, en orden descendente respectivamente.*

## 3) Reglas difusas

Reglas difusas lingüísticas para el sistema difuso con respecto a la calidad de los parámetros son las siguientes:

**R1:** Si el oxígeno disuelto es óptimo, el pH es óptimo y la temperatura es óptima entonces el estado del agente camarón es normal.

**R2:** Si el oxígeno disuelto es óptimo, el pH es óptimo y la temperatura es mala entonces el estado del agente camarón es tolerable.

**R3:** Si el oxígeno disuelto es óptimo, el pH es malo y la temperatura es óptima entonces el estado del agente camarón es tolerable.

**R4:** Si el oxígeno disuelto es malo, el pH es óptimo y la temperatura es óptima entonces el estado del agente camarón es tolerable.

**R5:** Si el oxígeno disuelto es malo, el pH es malo y la temperatura es óptima entonces el estado del agente camarón es malo.

**R6:** Si el oxígeno disuelto es malo, el pH es óptimo y la temperatura es mala entonces el estado del agente camarón es malo.

**R7:** Si el oxígeno disuelto es óptimo, el pH es malo y la temperatura es mala entonces el estado del agente camarón es malo.

**R8:** Si el oxígeno disuelto es malo, el pH es malo y la temperatura es mala entonces el agente camarón muere.

La regla número ocho en la que el agente camarón deja de existir en el ambiente no está contemplada en la simulación. El proceso mostrado en las reglas responde, a un esquema de modelado que permite manipular reglas de inferencia sobre conjuntos difusos, y que puede ser resumido como se presenta en la Figura 10.

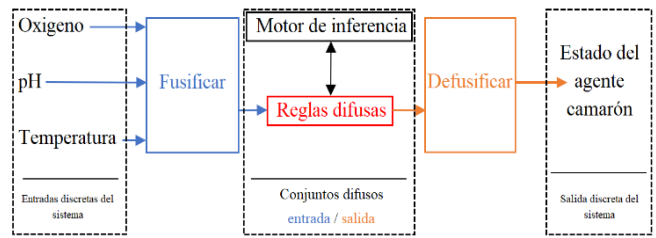

*Figura 10. Esquema del sistema difuso*

## 4) Sistema de razonamiento difuso

Una vez elegida la operación difusa e identificada las reglas, se puede obtener los valores fusificados en función de la calidad del ambiente, es decir, el grado de pertenencia. Luego estos valores son evaluados con las reglas difusas y se obtienen valores entre 0 y 1 con el cual se hará el proceso inverso, defusificar. Finalmente se toma un valor correspondiente a la escala del estado del agente camarón. Por ejemplo, si en la salida al sistema difuso se obtiene un valor de 0.2 de pertenencia, entonces su reflejo en el conjunto universo será estado malo. (Ver Figura 11)

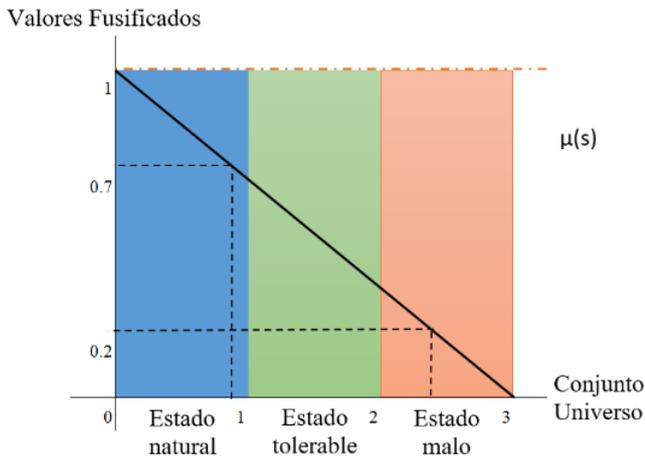

*Figura 11. Defusificación a partir de los valores obtenidos de las salidas del sistema difuso*

## III. ANALISIS DE RESULTADOS

Para el análisis de resultados se toma el error cuadrático medio y la desviación estándar como variables de comparación, entre los experimentos con respecto al crecimiento, principalmente para filtrar las mejores generaciones. Luego se establece cual es el experimento con la mejor configuración de combinaciones para una posible aplicación.

### A. Distribución del alimento

El riego del alimento se distribuye de dos maneras: Manual y automática. La distribución representada en el prototipo de simulación es la automática y esta tiene diferentes configuraciones. La disposición no es más que la ubicación del equipo, sobre o a los límites de la piscina del cultivo. Los acuicultores normalmente disponen de 1 a 3 equipos dependiendo del tamaño del cultivo, ubicados de manera equidistantes y centrados en la piscina de cultivo (Ver Figura 12).

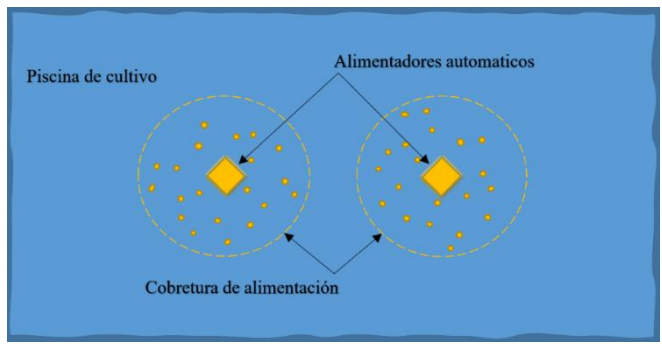

*Figura 12. Configuración de dos zonas de alimentación*

### B. Datos

Los datos son obtenidos del cultivo de camarones de la Isla Mondragón en Ecuador, estos registros son del control de crecimiento de 21 piscinas con alimentadores automáticos entre mayo del 2018 a marzo del 2020, con un total de 2215 registros por 22 columnas. Para el tratamiento de datos se tomó en cuenta solo 7 columnas, lo que nos deja una matriz de 2215x7. Los datos sin tratamiento estaban ordenados por semana e incluían el control de todas las piscinas. Entonces se procedió a ordenar por piscina, de tal manera que estuvieran todos los crecimientos en orden ascendente.

La base de expertos para la creación de las reglas de comportamiento de los agentes se obtuvo de entrevistas con profesionales del campo de la acuicultura, como lo fueron dos biólogos marinos, un ingeniero agrícola, un ingeniero industrial y un gerente productor de camaronera, además de la investigación que respalda al presente prototipo de simulación.

Una vez integrado todo el sistema, la simulación inicia sus pruebas para comparar la curva de los datos reales con los datos históricos. En la Figura 13, se presenta la regresión lineal de los datos históricos y de los datos de la simulación. Se simula con una configuración similar a la realidad, es decir con parámetros como la densidad de biomasa, la distribución del alimento y la cantidad de alimento.

Para llevar un control del crecimiento y determinar cuándo cosechar un cultivo de camarones, los expertos toman muestras de distintos puntos de la piscina semanalmente y promedian su tamaño. Una generación en la simulación se detiene cuando el tamaño promedio de los agentes camarón llega a 24gr. Este criterio de parada es un aproximado de los promedios cosechados en los cultivos reales.

El tratamiento de datos permite la ubicación de la simulación en términos de escala del tiempo. El criterio de parada por tamaño promedio permite hacer una relación entre la simulación y la realidad. Entonces 21 semanas en la realidad son aproximadamente 272 épocas en la simulación como se muestra en la Figura 13.

El error cuadrático medio, tanto en la simulación como en los datos reales, esta medido con el crecimiento en el tiempo y las épocas respectivamente con respecto a la regresión lineal de cada uno de estos conjuntos de datos. Para el caso de la desviación estándar también es calculada con los datos de crecimiento.

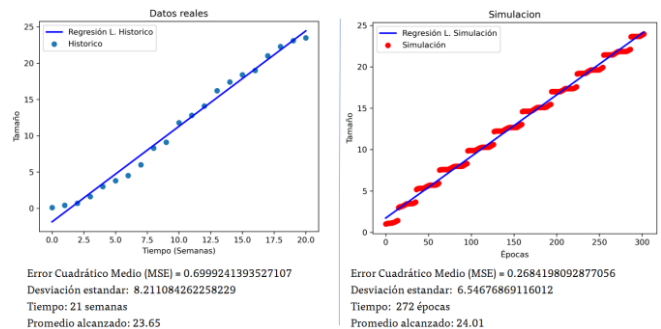

*Figura 13. Regresión lineal de los datos históricos y los datos simulados. Aproximación al tiempo total del cultivo*

## C. Experimentos

Se busca mejorar los procesos de producción, prueba de ello es la evolución de los métodos de alimentación. Como se mencionó en el apartado anterior, hay varias configuraciones en función del área del cultivo para la distribución del alimento.

El crecimiento del camarón va de 0.1 hasta 38 gr. Que son valores continuos, pero no se simula valores continuos, sino que se discretiza el tamaño en un rango de 1 a 38 gr. en pasos de 1gr. Un sembrío de camarones lo hacen a razón de millones de unidades, un cultivo contiene una media de 4'000.000 de ejemplares y en la simulación un agente representa a 8.000 camarones en la realidad esto es una escala muestral de 1:8.000, es decir, los 4 millones de camarones en la realidad son representados por 500 unidades en la simulación. La simulación experimenta con una, dos y tres zonas de alimentación para el análisis del crecimiento, revelando la cantidad de grupos formados de acuerdo con el tamaño del camarón, esto se representa con histogramas.

La cantidad de alimento es un porcentaje fijo de acuerdo con la escala. Presenta en la simulación dos metodologías de alimentación: Alimentación normal y alimentación alta, una difiere de la otra en un 20% adicional de alimento. La alimentación alta se usa cuando se quiere que el camarón tenga un crecimiento acelerado. Finalmente tenemos la densidad poblacional sobre un área y que está estrechamente relacionada con el tipo de cultivo si es semi intensivo o intensivo.

| No. Exp. | Disposición de la distribución del alimento | Modo de alimentación | Densidad pobalcional | No. Exp. | Disposición de la distribución del alimento | Modo de alimentación | Densidad pobalcional |
|---|---|---|---|---|---|---|---|
| 1 |  | Normal | Semi - Intensiva | 9 |  | Normal | Intensiva |
| 2 |  | Normal | Semi - Intensiva | 10 |  | Alta | Intensiva |
| 3 |  | Normal | Semi - Intensiva | 11 |  | Normal | Intensiva |
| 4 |  | Alta | Semi - Intensiva | 12 |  | Alta | Intensiva |
| 5 |  | Alta | Semi - Intensiva | 13 |  | Normal | Semi - Intensiva |
| 6 |  | Alta | Semi - Intensiva | 14 |  | Alta | Semi - Intensiva |
| 7 |  | Normal | Intensiva | 15 |  | Normal | Intensiva |
| 8 |  | Normal | Intensiva | 16 |  | Alta | Intensiva |

*Figura 14. Descripción de la combinación en los experimentos, la disposición del alimento va de una a tres zonas de alimentación más una distribución uniforme*

Con este preámbulo se pueden crear 16 configuraciones entre la disposición de la distribución del alimento, modo de alimentación y densidad poblacional, (Ver Figura 14), donde se considera las variaciones de:

- Disposición de la distribución del alimento, que va de una a tres zonas de alimentación como se maneja en la actualidad además de una última que es una distribución uniforme sobre toda la piscina.
- Modo de alimentación con dos valores, normal y alta.
- Densidad poblacional con dos valores, semi – intensiva e intensiva.

La simulación inicia con la configuración según la combinación de parámetros presentados en la Figura 14, posterior se genera la población según los cromosomas. Luego se levanta el ambiente en el que cobran vida los agentes e interactúan. Según la posición de los agentes camarones, este provoca un cambio en el ambiente. Dicho cambio es evaluado por lógica difusa y su resultado afecta directamente al estado del agente camarón.

En la simulación existen dos criterios de parada: El criterio de parada para la generación y el criterio de parada para la pre-experimentación. El criterio de parada de cada generación es el tamaño, es decir cuando el promedio del tamaño de los agentes camarón llega en este caso a 24 gr.

Cada una de las 16 configuraciones es usada como entrada al algoritmo genético en una etapa de pre-experimentación (Ver Figura 15). El criterio de parada de la pre-experimentación es un número fijo, y se determinó con experimentaciones aleatorias donde a partir de la 12va generación presentaron datos incoherentes o sin cambios, entonces se establecieron 10 generaciones como número fijo. Entre las 10 generaciones es seleccionado el experimento con menor error usando el MSE. Y esto da un total de 160 experimentos.

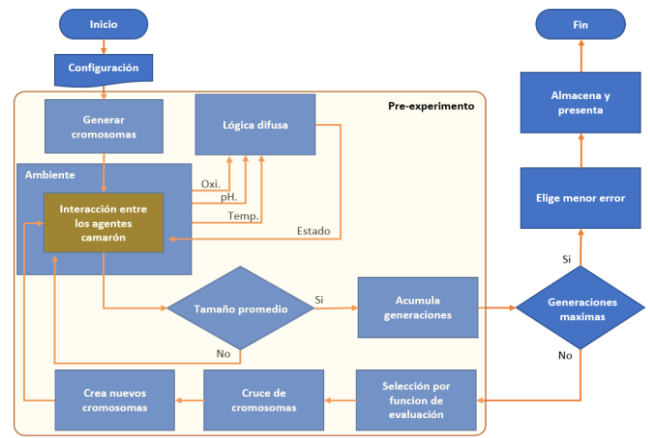

*Figura 15. Diagrama del funcionamiento general de la simulación. Etapa de pre-experimentación*

## D. Resultados

Se estableció una nomenclatura descrita en la Tabla 2, definiendo la combinación de los experimentos para facilitar la lectura de los resultados.

| Configuración | Nomenclatura | Descripción |
|---|---|---|
| Disposición de alimento | 3Z | 3 zonas de alimentación |
|  | 2Z | 2 zonas de alimentación |
|  | 1Z | 1 zona de alimentación |
|  | U | Alimentación Uniforme |
| Modo de alimentación | N | Normal |
|  | A | Alta |
| Densidad | I | Intensiva |
|  | S.I | Semi - Intensiva |

*Tabla 2. Nomenclatura de las configuraciones para la lectura de los resultados.*

Se presentan los 16 mejores resultados, en los que se agregó el número de épocas, el mínimo tamaño y el máximo tamaño obtenidos en cada experimento.

Como resultado de los 160 pre-experimentos, se presentan a continuación los histogramas de barras de la población en función de las agrupaciones por tamaño. (Ver de la Figura 16 a la Figura 31)

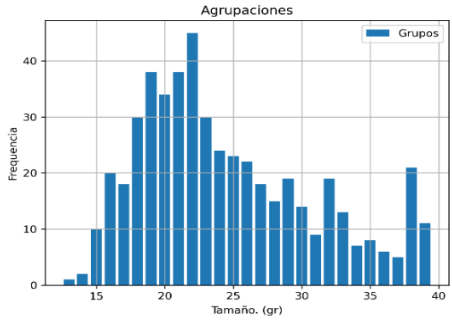

*Figura 16. Experimento 1: Configuración 3Z-N-I. Histogramas de la frecuencia de tamaños, para la muestra final en la simulación. Los histogramas se realizan con 40 cajas en el eje y*

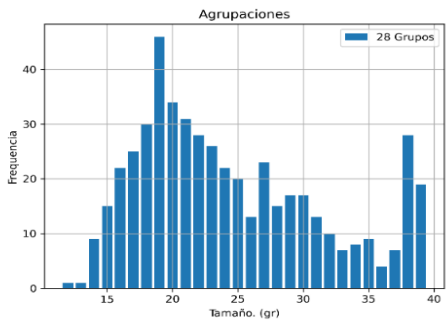

*Figura 17. Experimento 2: Configuración 2Z-N-I. Histogramas de la frecuencia de tamaños, para la muestra final en la simulación. Los histogramas se realizan con 40 cajas en el eje y*

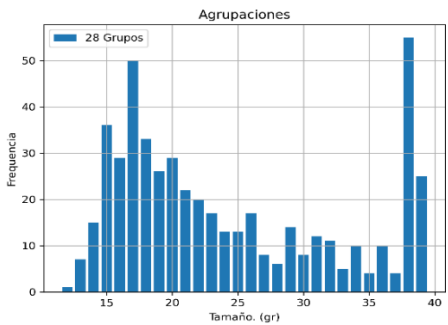

*Figura 18. Experimento 3: Configuración 1Z-N-I. Histogramas de la frecuencia de tamaños, para la muestra final en la simulación. Los histogramas se realizan con 40 cajas en el eje y*

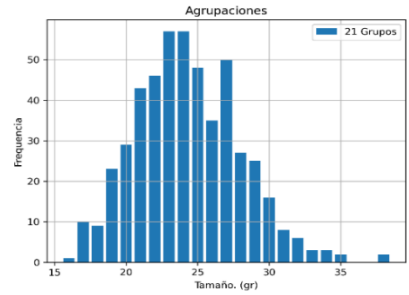

*Figura 19. Experimento 4: Configuración U-N-I. Histogramas de la frecuencia de tamaños, para la muestra final en la simulación. Los histogramas se realizan con 40 cajas en el eje y*

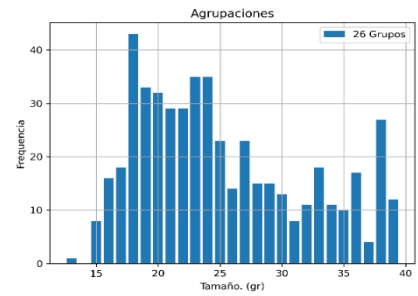

*Figura 20. Experimento 5: Configuración 3Z-A-I. Histogramas de la frecuencia de tamaños, para la muestra final en la simulación. Los histogramas se realizan con 40 cajas en el eje y*

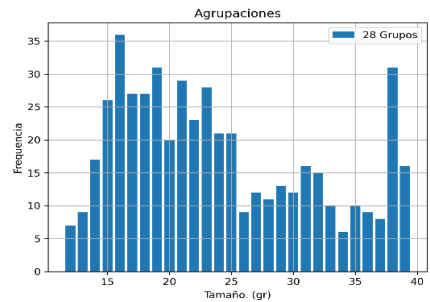

*Figura 21. Experimento 6: Configuración 2Z-A-I. Histogramas de la frecuencia de tamaños, para la muestra final en la simulación. Los histogramas se realizan con 40 cajas en el eje y*

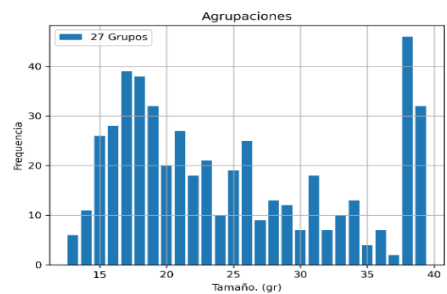

*Figura 22. Experimento 7: Configuración 1Z-A-I. Histogramas de la frecuencia de tamaños, para la muestra final en la simulación. Los histogramas se realizan con 40 cajas en el eje y*

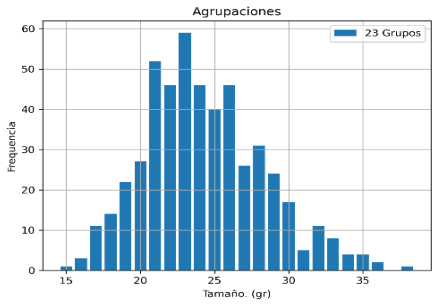

*Figura 23. Experimento 8: Configuración U-A-I. Histogramas de la frecuencia de tamaños, para la muestra final en la simulación. Los histogramas se realizan con 40 cajas en el eje y*

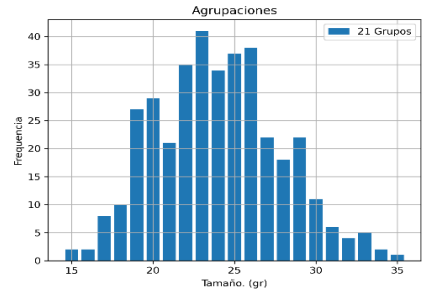

*Figura 27. Experimento 12: Configuración U-N-S.I. Histogramas de la frecuencia de tamaños, para la muestra final en la simulación. Los histogramas se realizan con 40 cajas en el eje y*

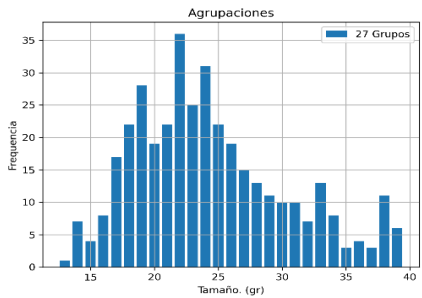

*Figura 24. Experimento 9: Configuración 3Z-N-S.I. Histogramas de la frecuencia de tamaños, para la muestra final en la simulación. Los histogramas se realizan con 40 cajas en el eje y*

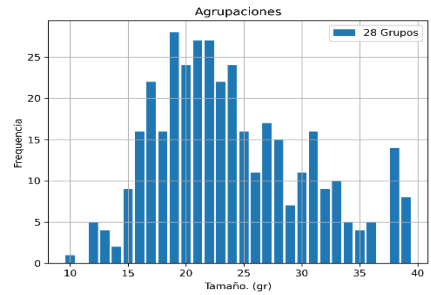

*Figura 28. Experimento 13: Configuración 3Z-A-S.I. Histogramas de la frecuencia de tamaños, para la muestra final en la simulación. Los histogramas se realizan con 40 cajas en el eje y*

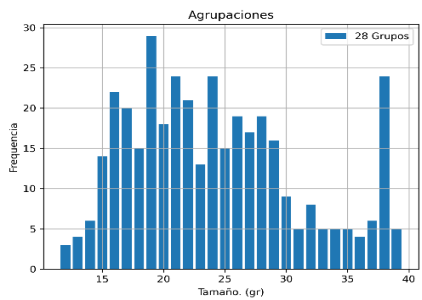

*Figura 25. Experimento 10: Configuración 2Z-N-S.I. Histogramas de la frecuencia de tamaños, para la muestra final en la simulación. Los histogramas se realizan con 40 cajas en el eje y*

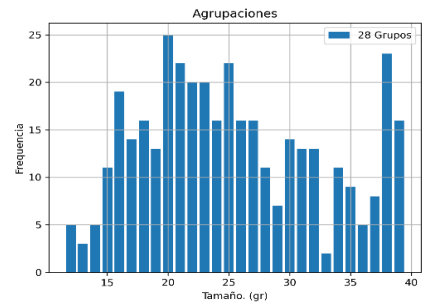

*Figura 29. Experimento 14: Configuración 2Z-A-S.I. Histogramas de la frecuencia de tamaños, para la muestra final en la simulación. Los histogramas se realizan con 40 cajas en el eje y*

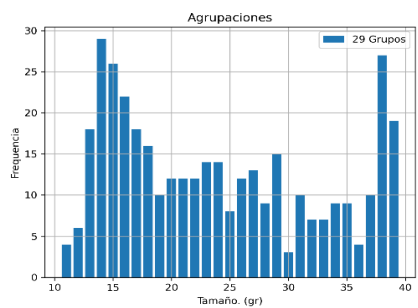

*Figura 26. Experimento 11: Configuración 1Z-N-S.I. Histogramas de la frecuencia de tamaños, para la muestra final en la simulación. Los histogramas se realizan con 40 cajas en el eje y*

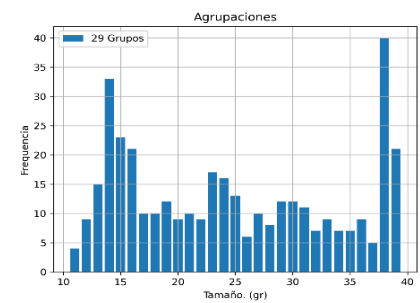

*Figura 30. Experimento 15: Configuración 1Z-A-S.I. Histogramas de la frecuencia de tamaños, para la muestra final en la simulación. Los histogramas se realizan con 40 cajas en el eje y*

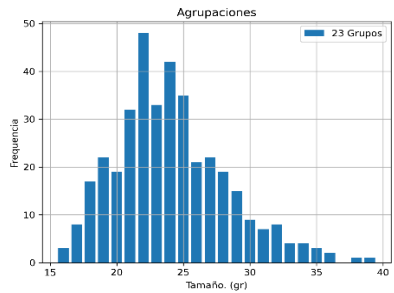

*Figura 31. Experimento 16: Configuración U-A-S.I. Histogramas de la frecuencia de tamaños, para la muestra final en la simulación. Los histogramas se realizan con 40 cajas en el eje y*

Analizando los histogramas, puede verse en la Figura 16, Figura 20, Figura 24 y Figura 28, donde el alimento se distribuye en tres zonas de alimentación, se nota un pico de crecimiento en los tamaños más grandes de la población. Lo mismo ocurre en los experimentos correspondientes a la Figura 16, Figura 17 y Figura 18 donde la cantidad de camarones de mayor tamaño aumenta conforme disminuye las zonas de alimentación, como se puede ver en la Figura 32. Esas agrupaciones en rojo corresponden a los agentes camarones que están más cerca del área de distribución del alimento.

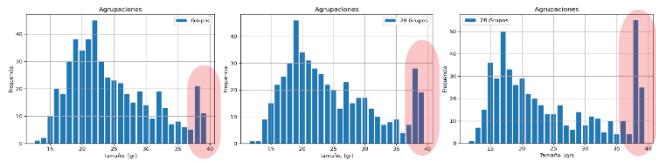

*Figura 32. Comparación de los experimentos 1, 2 y 3 donde se observa el aumento de la frecuencia en los tamaños en una de población. Corresponde a la Figura 16, Figura 17 y Figura 18 de izquierda a derecha. Las zonas de disposición del alimento corresponden respectivamente a 3, 2 y 1 zona de alimentación*

| No. Exp. | Config. | MSE | STD | Tamaño promedio | Tamaño Máx./Mín. | Épocas |
|---|---|---|---|---|---|---|
| 1 | 3Z-N-I | 0.22713411 | 7.04490504 | 24.598 | 40/13 | 289 |
| 2 | 2Z-N-I | 0.36926575 | 6.85140302 | 24.542 | 40/12 | 291 |
| 3 | 1Z-N-I | 0.29826879 | 7.07525894 | 24.358 | 40/12 | 315 |
| 4 | U-N-I | 0.48813656 | 6.83578311 | 24.334 | 38/16 | 233 |
| 5 | 3Z-A-I | 0.31213318 | 6.9950606 | 25.21 | 39/13 | 261 |
| 6 | 2Z-A-I | 0.36678564 | 6.74231586 | 24.024 | 39/12 | 263 |
| 7 | 1Z-A-I | 0.32614824 | 7.19412133 | 24.834 | 39/13 | 274 |
| 8 | U-A-I | 0.78948638 | 6.53729587 | 24.308 | 39/15 | 183 |
| 9 | 3Z-N-SI | 0.20660495 | 6.98038082 | 24.394 | 39/13 | 318 |
| 10 | 2Z-N-SI | 0.19848416 | 6.75116741 | 24.224 | 39/12 | 318 |
| 11 | 1Z-N-SI | 0.23882531 | 6.91226972 | 24.008 | 39/11 | 326 |
| 12 | U-N-SI | 0.40648533 | 6.63423219 | 24.0053 | 35/15 | 251 |
| 13 | 3Z-A-SI | 0.24654055 | 7.00549407 | 24.0106 | 39/10 | 277 |
| 14 | 2Z-A-SI | 0.24166166 | 7.42799731 | 25.424 | 39/12 | 283 |
| 15 | 1Z-A-SI | 0.3947841 | 6.88623758 | 24.8186 | 39/11 | 288 |
| 16 | U-A-SI | 0.53703271 | 6.7284968 | 24.08 | 39/16 | 205 |

*Tabla 3. Resultados obtenidos, la nomenclatura de la columna de configuración esta descrita en la Tabla 2*

La tasa de error en este punto del análisis ya no es un criterio de selección. Hay que recordar que cada configuración tuvo 10 experimentos y el experimento seleccionado es el que tiene menor tasa de error. Para este análisis el criterio de selección está dirigido a las épocas y a la desviación estándar.

Las épocas son las acciones entre los cambios de un estado S a un estado S+1 en la simulación, en términos informáticos son los estados o ciclos de una sentencia repetitiva y en esta investigación se asocia el total de estas épocas en la simulación al tiempo total del cultivo en la realidad. Con esta base se juzga como las mejores, a las configuraciones que tengan menor número de épocas y la desviación estándar más cerrada.

Ahora el siguiente análisis está enfocado en la configuración de densidad que es cuando se tiene dos configuraciones en base al tipo de cultivo, si es intensivo o semi intensivo. Se dice que un cultivo es semi intensivo cuando la densidad poblacional es inferior a 200.000 animales por hectáreas e intensivo cuando es superior [9].

Se pueden dividir los experimentos en 2 grupos, un grupo de densidad intensiva (I) y otro grupo de densidad semi intensiva (S.I), visto de esta manera se puede observar en la Tabla 3 que los experimentos de densidad semi intensiva demoran más, es decir tienen mayor cantidad de épocas, mientras que los experimentos de densidad intensiva llegan al criterio de parada más rápido. Esto se debe al comportamiento emergente de enjambre o *"flocking"*. Cuando un alimento se esparce no solo un agente lo detecta, sino todos los agentes que están alrededor de él. Entonces cuando es semi intensiva hay menor población distribuida, hay mayores distancias o vacíos entre los grupos que se alimentan, por ende, más tiempo en encontrar el siguiente alimento. A diferencia de la densidad intensiva donde hay una mayor población y una distribución equilibrada de esta población en el ambiente.

De los grupos, intensiva (I) y semi - intensiva (SI) los vemos desde la perspectiva de la configuración "modo de alimentación" que son alta (A) y normal (N). Como ya se había mencionado antes, la diferencia entre estos modos consiste en si se requiere un crecimiento acelerado o no; y esto se confirma en los resultados presentados en la Tabla 3 donde se observa que los experimentos de alimentación alta alcanzan el criterio de parada en menos épocas que los experimentos de alimentación normal.

Se forman 4 grupos de 4 experimentos, solo que en esta ocasión en función de la configuración de la disposición del alimento. Se observa que los experimentos 3, 7, 11 y 15, donde la configuración por disposición de alimento es de solo una zona de alimentación tienen el mayor número de épocas entre todos los experimentos, este fenómeno tiene sentido ya que el alimento está distribuido en una sola zona de la piscina, entonces para que el agente camarón llegue al criterio de parada le toma más interacciones para crecer. Esto se debe a que el agente camarón que esta fuera del alcance de la zona de alimentación le toma más tiempo desarrollarse por su limitada alimentación. Entre las configuraciones de dos y tres zonas de alimentación no existen muchas variantes y esto se acerca a la realidad ya que son las más utilizadas en los cultivos camaroneros.

Para concluir se dirige la atención a la configuración de disposición del alimento, específicamente en la distribución uniforme. Dicha configuración es el caso extremo de la distribución del alimento y causa del desarrollo de esta investigación. En la Tabla 3 se observa los experimentos 4, 8, 12 y 16 que tienen de hecho el error más alto entre todos los experimentos, pero hay que recordar que ese ya no es el criterio de selección en este punto del análisis, sin embargo, esto tiene una explicación, retornando a los histogramas se observa que el comportamiento en la configuración de distribución uniforme del agente camarón es totalmente diferente.

El agente camarón no recorre mayor distancia para obtener el alimento, baja el nivel de competencia por el alimento, se reduce la población de agentes camarones que se sobre desarrollan (Ver Figura 32) y dejan sin alimento al agente pequeño, los parámetros físicos químicos del contenedor son estables por lo que el agente no entra en conflicto por si alimentarse o no, si el alimento está en la misma área del ambiente de mala calidad. Cuando todo esto ocurre, el número de interacciones para alcanzar el criterio de parada se reduce drásticamente en cualquiera de las configuraciones. No solo eso sino también se reduce las agrupaciones por tamaño lo que significa que su promedio es menos disperso. (Ver Figura 33)

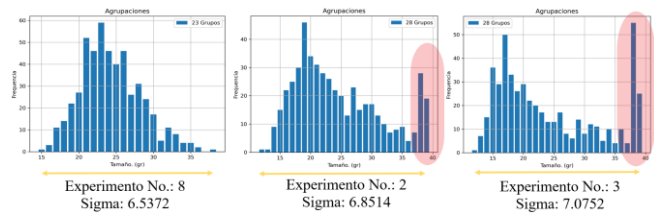

*Figura 33. Comparación de la configuración de distribución uniforme contra distribución de 2 y 1 zona de alimentación. Correspondientes a los experimentos 8, 2 y 3 de izquierda a derecha respectivamente*

## IV. Conclusiones

Al inicio de la investigación se decidió implementar en el prototipo tres parámetros físicos químicos para simular la calidad del ambiente, en el desarrollo se llegó a considerar que eran insuficientes, pero ya en la implementación y con pruebas preliminares en lógica difusa la toma de decisiones de los agentes cumplió con el comportamiento esperado. Cabe destacar que lógica difusa puede manejar como entrada todas las variables que intervienen en la calidad del ambiente, con más reglas, más posibilidades de cubrir una extensa variedad de comportamientos.

La configuración de una sola zona de alimentación no es común en la práctica, a diferencia de las configuraciones de dos a tres zonas de alimentación. Con esto se ratifica y justifica la razón del no uso de una sola zona de alimentación. Existe una excepción a la regla, que es cuando las dimensiones de la piscina son muy pequeñas, es decir menor a 1ha.

De cada grupo formado para facilitar el análisis, se observa que cada configuración tiene su mejor opción. En base a esto se concluye que la mejor configuración es: Una distribución uniforme con una alimentación alta y una densidad intensiva. Se confirma esto en el experimento número 8 donde se llega al criterio de parada en el menor número de épocas. (Ver Figura 34)

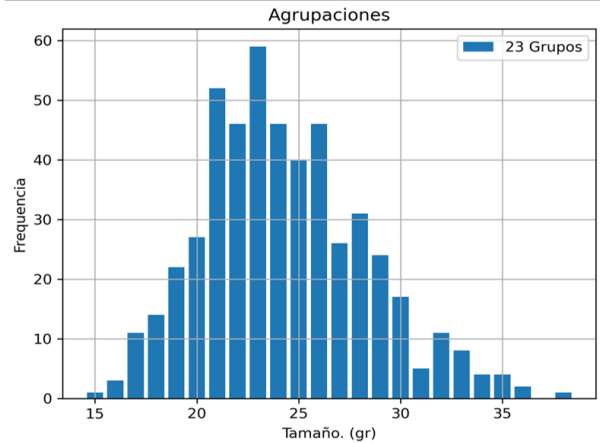

*Figura 34. Experimento 8 con el menor número de épocas*

Promediando las épocas en las configuraciones de dos y tres zonas de alimentación se tiene un valor de 287,5 épocas, que representan a 22 semanas de cultivo. Con lo que se dice que el experimento numero 8 con 183 épocas corresponde en la simulación a 14,0034 semanas. Es decir 7,9966 semanas menos que la producción normal.

Como resultado de las simulaciones realizadas, puede considerarse que si se desarrolla una tecnología que permita una distribución uniforme del alimento, deberá estudiarse si los costos que implican la implementación de esta técnica de distribución del alimento, se compensan con el 16.41% de mejora simulada que se obtendría en la producción.

La combinación de algoritmos genéticos, lógica difusa y multi-agentes hizo posible que la presente investigación abarcara más allá de los objetivos establecidos inicialmente, no solo llegando a una comparación del prototipo de una realidad simulada con los registros históricos, sino se llegó a comparar escenarios reales contra una posibilidad no estudiada en la realidad, por su alto grado de dificultad como lo es la distribución uniforme del alimento.

El camarón como entidad posee características tanto individuales como colectivas, esto convierte en un desafío lograr imitar su comportamiento. El diseño de esta entidad contiene características bases de su comportamiento colectivo, como alimentación, factor de crecimiento, estado y desplazamiento. El diseño de las características individuales es un reto que se puede asumir en futuras investigaciones.